%% file: Ji-et-al.tex
\documentclass[graybox]{svmult}

\usepackage{mathptmx}       
\usepackage{helvet}         
\usepackage{courier}        
\usepackage{type1cm}        
\usepackage{graphics}
\usepackage{makeidx}         
\usepackage{graphicx}        
\usepackage{multicol}        
\usepackage[bottom]{footmisc}

\usepackage{subcaption}
\usepackage{caption}

\usepackage{amsmath}
\usepackage{amssymb}
\usepackage{graphicx}
\usepackage{graphicx, wrapfig, setspace, booktabs}

\makeindex             

\begin{document}

\title*{Emotion pattern detection on facial videos using functional statistics}
\author{Rongjiao Ji, Alessandra Micheletti, Natasa Krklec Jerinkic, Zoranka Desnica}
\institute{Rongjiao Ji, Alessandra Micheletti \at Universit\'a degli Studi di Milano, \email{rongjiao.ji@unimi.it, alessandra.micheletti@unimi.it}
\and Natasa Krklec Jerinkic \at University of Novi Sad,  \email{natasa.krklec@dmi.uns.ac.rs}
\and Zoranka Desnica \at 3Lateral DOO, \email{zoranka.desnica@3lateral.com} }
%
%
\maketitle

\abstract{There is an increasing scientific interest in automatically analysing and understanding human behavior, with particular reference to the evolution of facial expressions and the recognition of the corresponding emotions. In this paper we propose a technique based on Functional ANOVA to extract significant patterns of face muscles movements, in order to identify the emotions expressed by actors in recorded videos.  We determine if there are time-related differences on expressions among emotional groups by using a functional F-test. Such results are the first step towards the construction of a reliable automatic emotion recognition system \footnote{This work was funded by European Union’s Horizon 2020 research and innovation programme under the Marie Skłodowska Curie grant agreement No 812912 for the project BIGMATH.}}
\abstract{\emph{C'\`e un crescente interesse scientifico nell'analizzare e intepretare automaticamente il comportamento umano, soprattutto rispetto all'evoluzione delle espressioni del volto e al riconoscimento delle corrispondenti emozioni espresse. In questo lavoro proponiamo una tecnica, basata sull'ANOVA Funzionale per estrarre pattern significativi dei movimenti dei muscoli facciali, al fine di identificare le emozioni espresse da alcuni attori in video registrati.  In particolare determiniamo se, in istanti specifici, ci siano differenze nell'evoluzione delle espressioni fra diversi gruppi di emozioni, applicando un F-test funzionale. Questi risultati sono il primo passo verso la costruzione di un sistema affidabile per il riconoscimento automatico delle emozioni.}}


\section{Introduction}
\input{1_introduction}

\section{The RAVDESS Dataset}
\input{2_data}

\section{Functional Statistical Methods}
\input{3_methods}

\section{Results}
\input{4_results}

\bibliographystyle{plain}
\bibliography{bibliography.bib}

\end{document}

%% file: 1_introduction.tex
The study of human facial expressions and emotions never stops in our daily life while we communicate with others. Following the increased interest in automatic facial behavior analysis and understanding, the need of a semantic interpretation of the evolution of facial expressions and of human emotions has become of interest in recent years \cite{Fridlun2014}. In this paper, based on a work cooperated with the Serbian company 3Lateral, which has special expertise on building visual styles and designs in animation movies, we want to explore functional statistical instruments to identify the emotions while analyzing the expressions through recorded videos of human faces. The final aim of this research is to use this information to better and more realistically establish virtual digital characters, able to interact autonomously with real humans. 

The data that we consider are multivariate longitudinal data, showing the evolution in time of different face muscles contraction. Functional Data Analysis (FDA) offers the possibility to analyze the entire expression evolution process over time and to gain detailed and in-depth insight into the analysis of emotion patterns. The basic idea in functional data analysis is that the measured data are noisy observations coming from a smooth function.  Ramsay and Silverman \cite{ramsay1997functional} describe the main features of FDA, that can be used to perform exploratory, confirmatory or predictive data analysis.  Ullah and Finch \cite{UllahShahid2013} published a systematic review on the applications of functional data analysis, where they included all areas where FDA was applied.


In our application, Functional ANOVA can be used to determine if there are time-related differences between emotion groups by using a functional F-test \cite{Dannenmaier2020}. Functional ANOVA yields the possibility to determine if a functional response can be described by scalar or functional variables.

The structure of this paper goes as follows. In Section 2 we briefly describe the RAVDESS dataset from where the expression data of interest is extracted. Section 3 includes some methods of functional data analysis that we implemented in our application, and in Section 4 our results are presented.

%% file: 2_data.tex
RAVDESS (Ryerson Audio-Visual Database of Emotional Speech and Song) \cite{livingstone2018ryerson} fits our needs for studying the human expression evolution and emotion identification, as it contains 24 professional actors (12 female, 12 male) to offer the performance with good quality and natural behavior under the emotions: calm, happy, sad, angry, fearful, disgusted and surprised. Also a neutral performance is available for each actor. The actors are vocalizing one lexically-matched statement in a neutral North American accent (“Kids are talking by the door”).

To avoid being lost in the difference of individual facial appearances, when analyzing the expressions and emotions, researchers mostly focus on the movements of individual facial muscles which are encoded by the Facial Action Coding System (FACS) \cite{ekman1997face}. FACS is a common standard to systematically categorize the physical expression of emotions, extracting the geometrical features of the faces and then producing temporal profiles of each facial movement. Such movements, corresponding to contraction of specific muscles of the face, are called \emph{action unit (AU)}.
As action units are independent of any interpretation, they can be used for any higher-order decision-making process including recognition of basic emotions. Following the FACS rules, OpenFace \cite{amos2016openface}, an open-source software, is capable of recognizing and extracting facial action unit from facial images or videos. We applied OpenFace to extract the engagement degrees of action units for the videos in RAVDESS. The extracted action units include 17 functions for each video, taking values in $[0,5]$, sampled in about 110 time points (which is also the number of frames in each video) varying around 110. 





%% file: 3_methods.tex



We will represent the action units evolution recorded on each video as a multivariate time series $\mathbf{Y}(t)= {(Y_{1}(t), \dots,Y_{d}(t), \dots, Y_{D}(t)), t \in [0, T]}$ containing a set of $D$ univariate longitudinal functions ($D=17$ in our case), each defined on the finite interval $[0, T], 0 < T < +\infty.$
The observation of $\mathbf{Y}$ on our sample of videos provides the set ${\mathbf{Y_1},\dots, \mathbf{Y_n}}$ of multivariate curves, that we represent as multivariate functional data.



It is essential to align the action units functions into a common registered internal timeline that follows the same pronunciation speed, to control the influence of the specific pronounced sentence and to detangle it from the influence of the emotions. Therefore, we need to isolate the phase variability of the action units curves, but keeping, at the same time, the amplitude-phase unchanged to maintain the information of the intensity level of the action units. 

The phase variation is normally represented by a random change of time scale, which is mostly a non-linear transformation. 
We use the warping functions $T_i: [0,T] \rightarrow [0,T], i = 1, \dots, n$, assuming that they are increasing functions independent of amplitude variation. They map unregistered chronological time $t_i^*$ to registered internal time $t$ so that $T_i^{-1}(t_i^*) = t$, with $E[T_i(t)] = t$. The observed time-warped curves, represented through a Karhunen-Loeve expansion based on a functional basis $\mathbf{f}_{j,d}$, are
$$\tilde{Y}_{i,d}(t) = Y_{i,d}(T_i^{-1}(t_i^*)) = \mu_{d}(T_i^{-1}(t_i^*)) + \sum\limits_{j \geq 1} C_j\mathbf{f}_{j,d} (T_i^{-1}(t_i^*)),$$




We used a spline basis and followed the principal components based registration method with a generative process \cite{wrobel2019register}, whose codes are available in the R package "registr" \cite{wrobel2018register}. 



  Using the registered curves representing the AUs evolution in each video, we then investigated if there exist patterns which could discriminate the different emotions, using a Functional ANOVA model.
  
  Let $y_{k,g}(t)$ be the evolution of one specific action unit in the video $k \in \{1,\dots,K\}$ (in our case $K=48$) for emotion $g\in \{1,\dots,7\}$. We can assume that
\begin{equation}
y_{k,g}(t) = \mu_0(t) + \alpha_{g}(t) + \epsilon_{k,g}(t),
\label{flm:indiv}
\end{equation}
where $\mu_0(t)$ is the grand mean function due to the pronounced sentence and to the actor, independent from all emotions. The term $\alpha_{g}(t)$ is the specific effect on the considered action unit of emotion $g$, while $\epsilon_{k,g}(t)$ represents the unexplained zero mean variation, specific of the $k$-th video within emotion group $g$. To be able to identify them uniquely, we require that they satisfy the constraint
$\sum\limits_{g=1}^7 \alpha_{g}(t) = 0, \forall t.$

By grouping the videos representing the same emotion, we can define a $8K \times 8$ design matrix $\mathbf{Z}$ for this model, with suitable 0 and 1 entries, as described in \cite[Section 9.2]{ramsay1997functional}, and rewrite Equation \ref{flm:indiv} in  matrix form:
$\mathbf{y}=\mathbf{Z}\mathbf{\beta} + \mathbf{\epsilon},$ where $\mathbf{\beta}=[\mu_0(t),\alpha_1(t),\dots ,\alpha_7(t)]^T$.

To estimate the parameters we use the functional least squares fitting criterion 
\begin{equation}
   \hat{\mathbf{\beta}}(t) = \arg \min\limits_{\mathbf{\beta}} \sum\limits_{g=1}^8 \sum\limits_{k=1}^K \int_0^T [y_{k,g} (t) - \langle z_{k,g},\mathbf{\beta}(t) \rangle]^2 dt,
\end{equation}
subject to the constraint 
$0 = \sum\limits_{i = 1}^7 \alpha_i(t) = \sum\limits_{j = 2}^8 \beta_j(t),\ \  \forall t$.

In order to investigate which emotions are significantly influencing the change of the action units patterns, for each emotion $\tilde{g}$ and for each action unit we test the null hypothesis
$ H_0: \alpha_{\tilde{g}}(t) = 0.$


Similarly to the classical univariate ANOVA model, the statistics used to test $H_0$ is






\begin{equation*}
    FRATIO(t) = \frac{MSR(t)}{MSE(t)} 
\end{equation*}
that under $H_0$ has an F distribution with suitable degrees of freedom.

%% file: 4_results.tex

As mentioned before, we first aligned the curves by separating the amplitude and phase variability. We choose to align the curves by AU25, which represents the lip movement, and then we adjusted the time frames of the other AUs according to this rescaling.

\begin{figure}[h]
\centering

\begin{subfigure}{.32\textwidth}
  \centering
  \includegraphics[width=\linewidth]{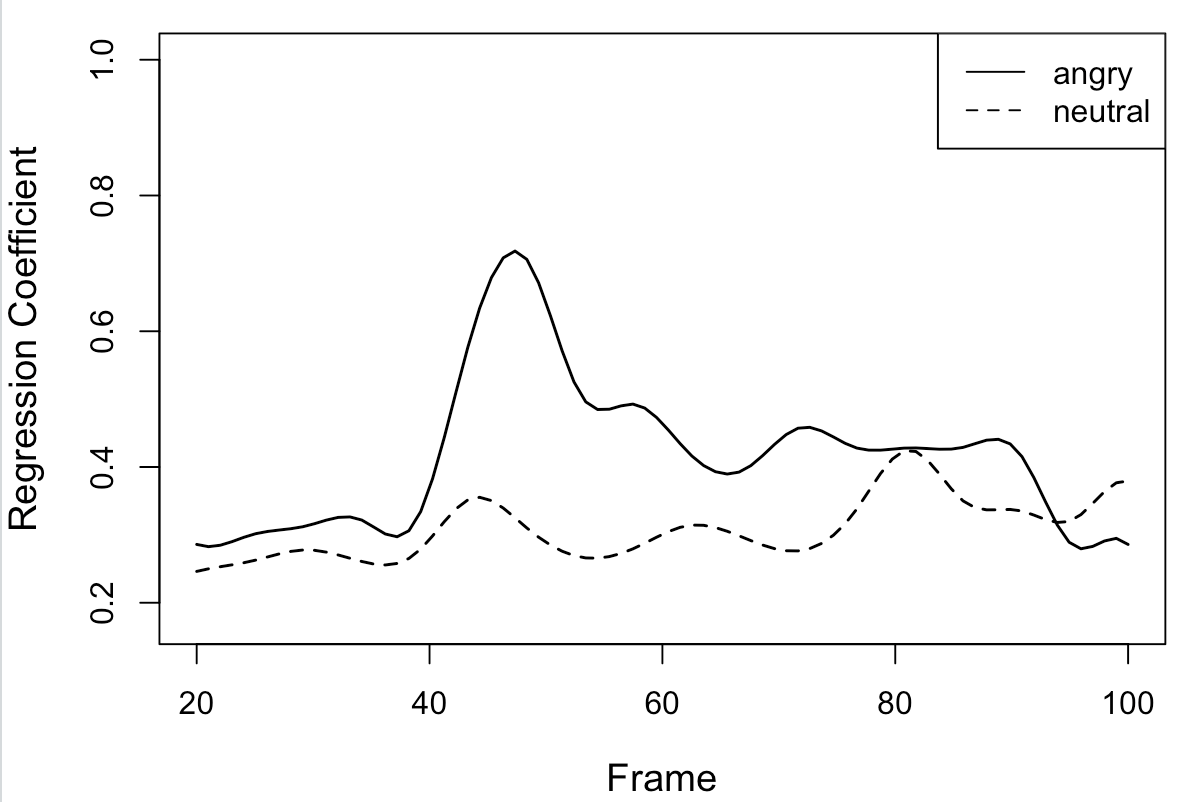}  
  \caption{AU07.}
  \label{fig:07f}
\end{subfigure}
\begin{subfigure}{.32\textwidth}
  \centering
  \includegraphics[width=\linewidth]{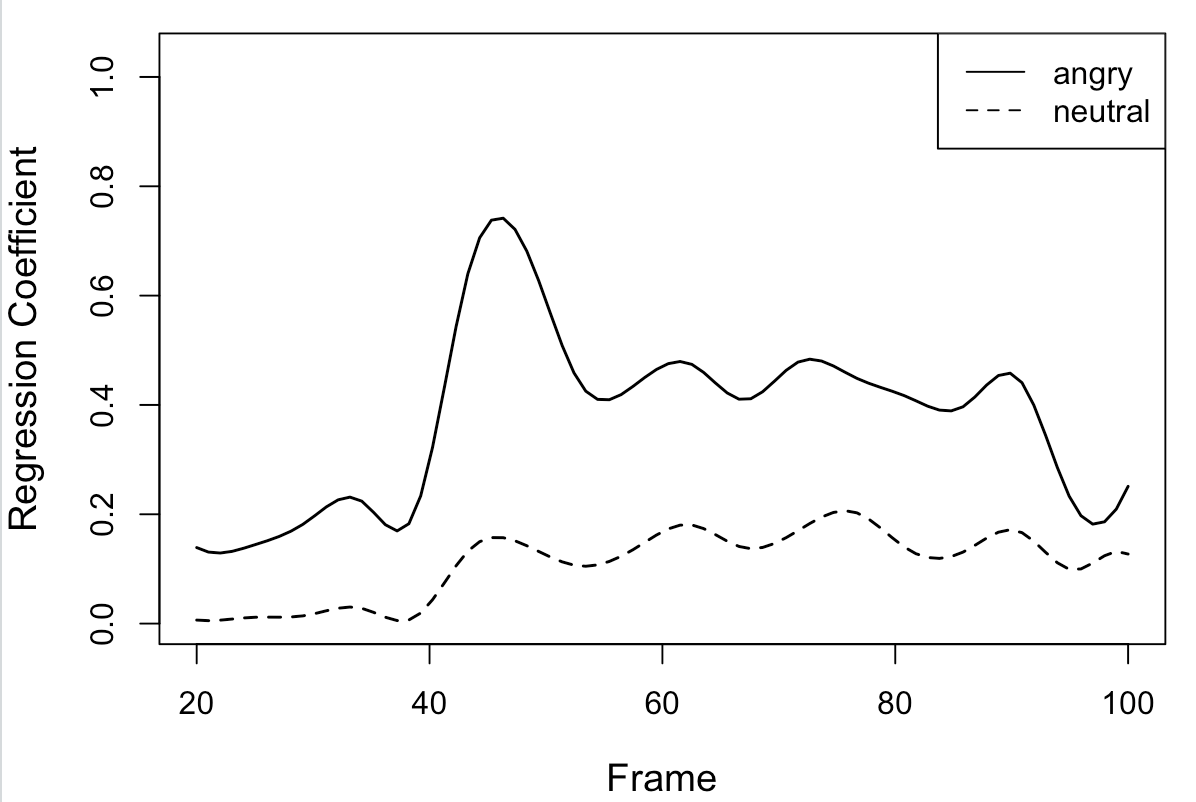}  
  \caption{AU10.}
  \label{fig:10f}
\end{subfigure}
\begin{subfigure}{.32\textwidth}
  \centering
  \includegraphics[width=\linewidth]{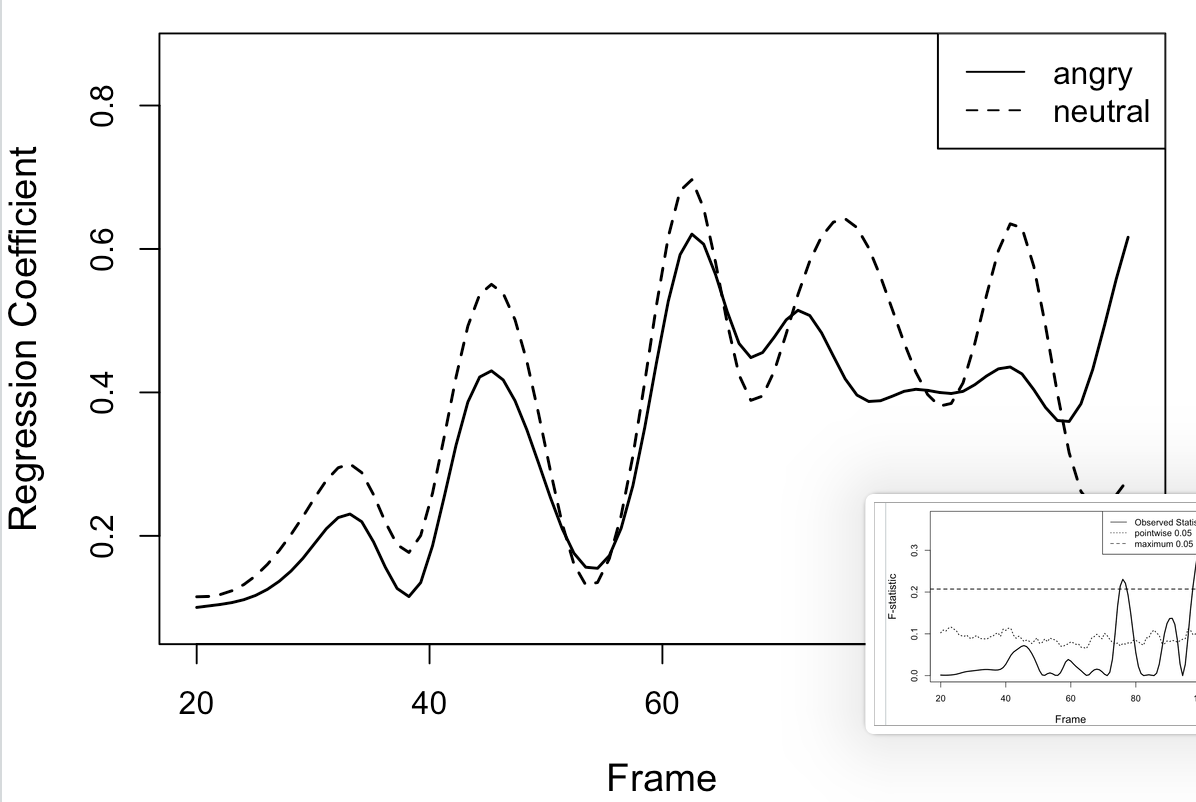}  
  \caption{AU26.}
  \label{fig:26f}
\end{subfigure}

\begin{subfigure}{.32\textwidth}
  \centering
  \includegraphics[width=\linewidth]{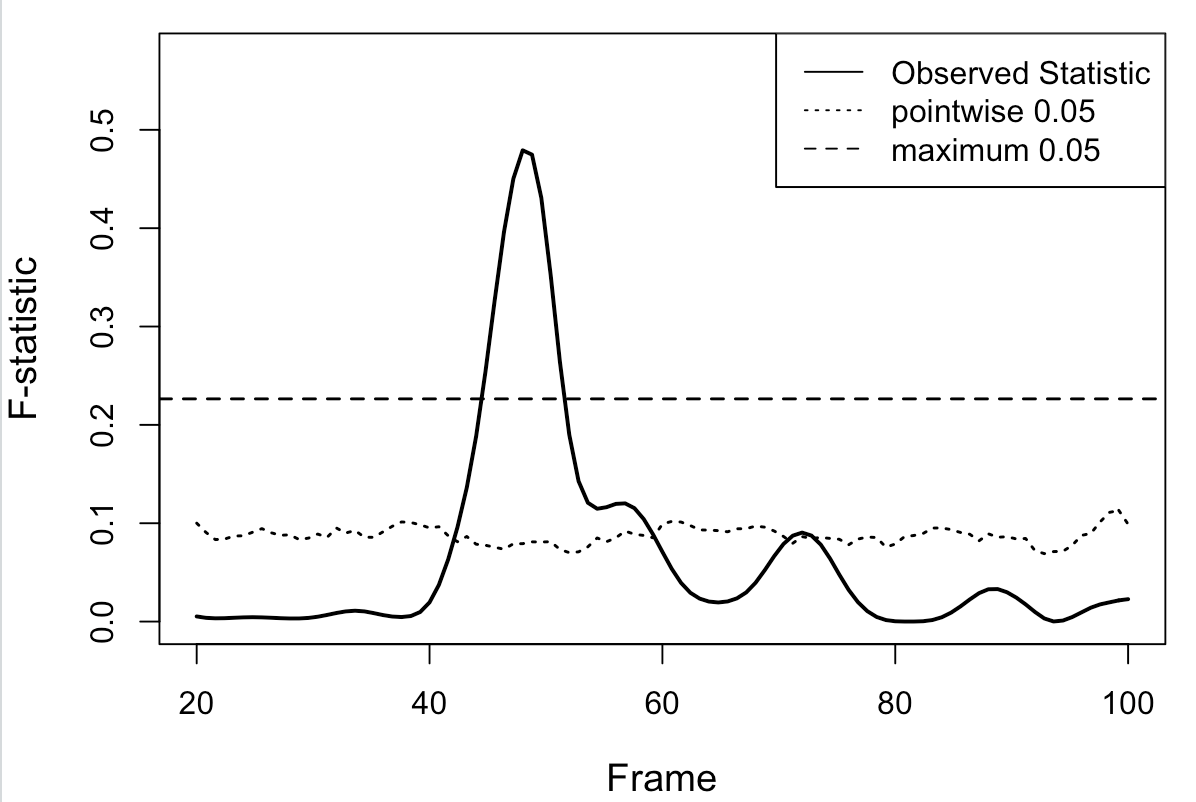}  
  \caption{AU07.}
  \label{fig:07c}
\end{subfigure}
\begin{subfigure}{.32\textwidth}
  \centering
  \includegraphics[width=\linewidth]{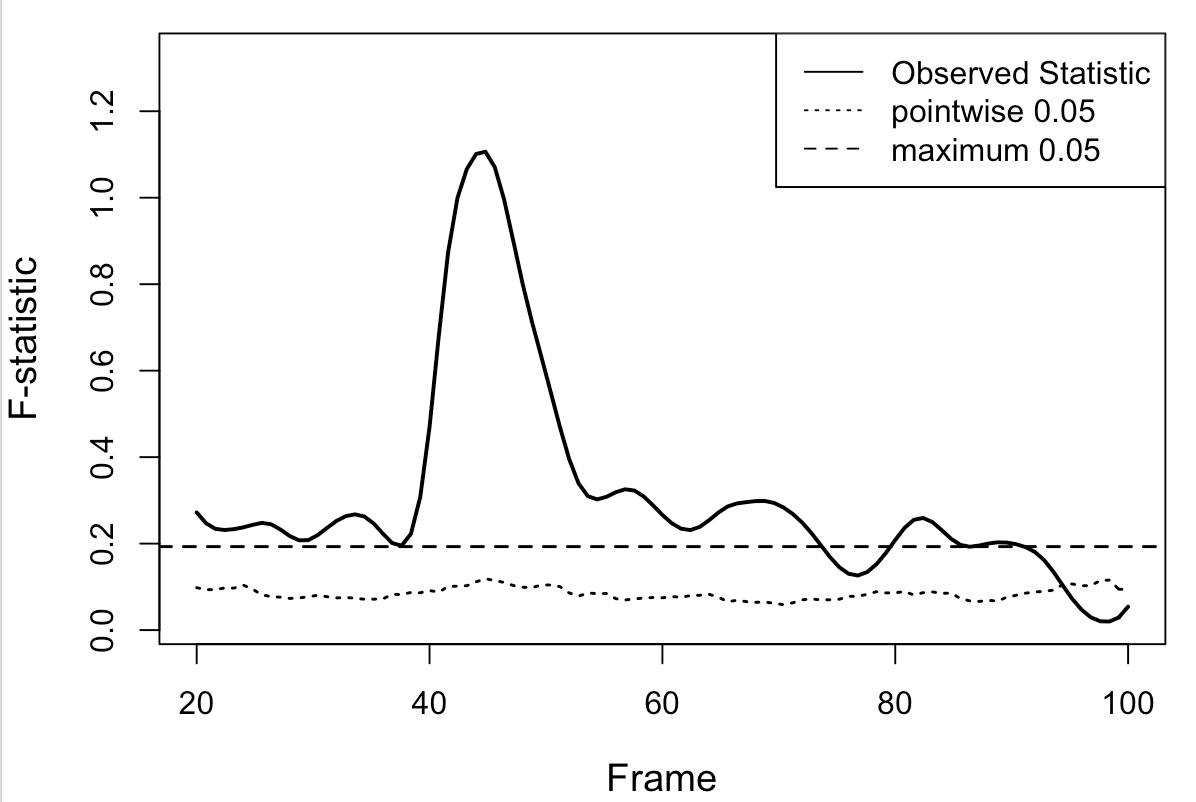}  
  \caption{AU10.}
  \label{fig:10c}
\end{subfigure}
\begin{subfigure}{.32\textwidth}
  \centering
  \includegraphics[width=\linewidth]{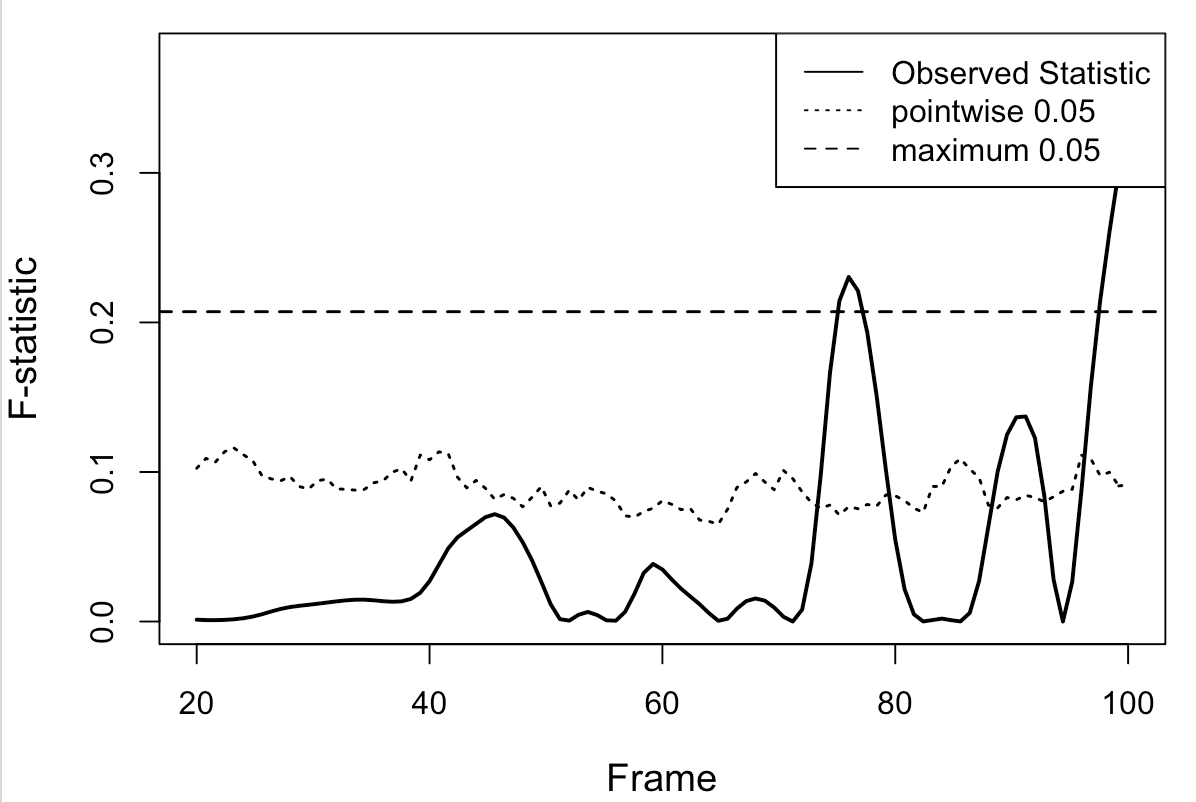}  
  \caption{AU26.}
  \label{fig:26c}
\end{subfigure}

\caption{The functional coefficients  of action units 07 (Lid Tightener), 10 (Upper Lip Raiser) and 26 (Jaw Drop) under neutral and angry emotion and the corresponding F-test results}
\label{fig:regression}
\end{figure}

We then applied the F-test described in the previous section to detect, for each emotion, which AUs have a mean behaviour significantly different from the neutral performance and in which time period during the videos.
In Figure \ref{fig:regression} we illustrate the results for emotion angry, as an example.

\begin{figure}[h]
    \centering
    \includegraphics[width=0.8\linewidth]{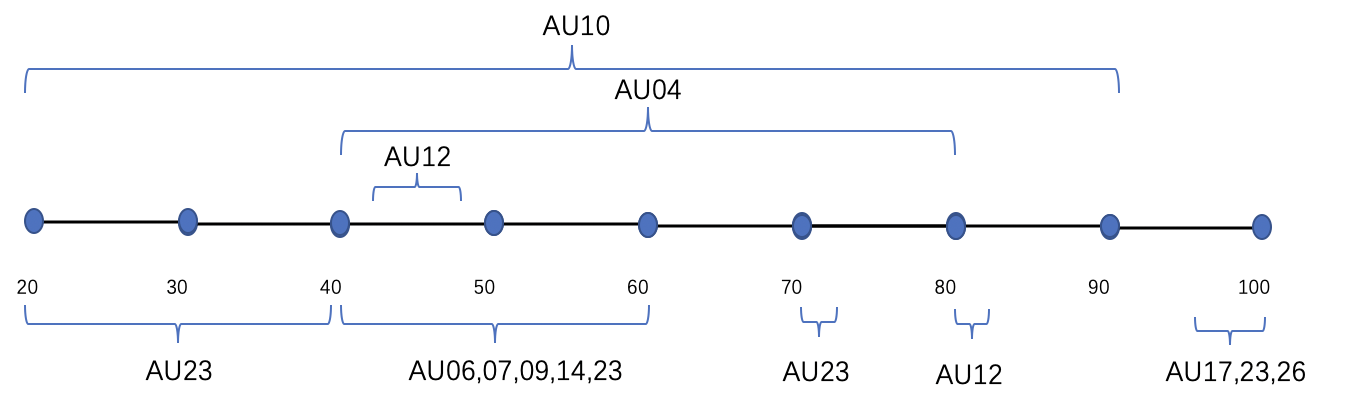}  
    \caption{Which and where AU values are affected significantly by angry emotion }
    \label{fig:angry.timeline}
\end{figure}

The first row of Figure \ref{fig:regression} illustrates the estimated mean $\mu_0(t)$ (neutral emotion) and the angry emotion effects for three action units. The second row displays the observed F-statistics curves together with the pointwise and maximum 95\% significance level for the F-distribution in the dashed and horizontal dotted lines respectively. Thus  when the observed F-statistics is higher than the critical level lines, the emotion has a significant effect on the AU's pattern. We found in general three main situations of influence of one emotion on expression evolution: 1. locally strengthening (Figure \ref{fig:07c}: AU07 in frame range 45 to 55) 2. locally inhibiting (Figure \ref{fig:26c}: AU26 in frame range 70 to 90) 3. globally strengthening (Figure \ref{fig:10c}: AU10 in almost the whole time). Further, we pointed out the time zones of significant effects of the angry emotion on the action units in Figure \ref{fig:angry.timeline}, which is beneficial to understand and detect dynamically when and how the facial muscles contractions differ from the baseline.

\begin{table}[htbp]
  \centering
  \small
  \caption{Emotions with corresponding significant action units}
    \begin{tabular}{c|c}
    \hline
    \hline
    Emotions & Related Action Units\\
    \hline
    Calm  & $06, 07, 10, 12, 14, 23$\\
    Happy & $01, 06, 07, 10, 12, 14, 17, 23, 25, 26$ \\
    Sad   & $04, 06, 10, 14, 17, 20, 23, 25$ \\
    Angry & $04, 06, 07, 09, 10, 12, 14, 17, 23, 26$ \\
    Fearful & $04, 09, 10, 12, 14, 15, 17, 23, 25, 26$ \\
    Disgust & $04, 06, 07, 09, 10, 12, 14, 17, 23, 25, 26$ \\
    Surprised & $06, 09, 10, 12, 14, 15, 17, 23, 25, 26, 45$ \\
    \hline
    \hline
    \end{tabular}%
  \label{tab:emoau} 
\end{table}

Table \ref{tab:emoau} summarizes for each emotion of interest the related action units that show significant changes from the neutral case  for our videos dataset. Similarly to the example of angry, we found that for happy and disgust emotions more action units have the globally strengthening effect on a large time range. Sad emotion sometimes affects the action units to be more constant than in neutral case. Emotion Fearful has more influence on upper half face (brows, eye lids and nose), while emotion calm is more related with the center of the face (Cheek Raiser, Lid Tightener and Lip Corner Puller). Surprised emotion is the only emotion where AU45 is significantly influenced. 

As a conclusion, our results can be joined in a multivariate setting and exploited to build a classifier able to automatically recognize the emotions. This task is left to subsequent works.

%% file: Ji-et-al.bbl
\begin{thebibliography}{1}

\bibitem{amos2016openface}
B.~Amos, L.~Bartosz, and M.~Satyanarayanan.
\newblock Openface: A general-purpose face recognition library with mobile
  applications.
\newblock Technical report, CMU-CS-16-118, CMU School of Computer Science,
  2016.
\newblock https://cmusatyalab.github.io/openface/.

\bibitem{Dannenmaier2020}
J.~Dannenmaier, C.~Kaltenbach, T~Kölle, and G.~Krischak.
\newblock Application of functional data analysis to explore movements:
  walking, running and jumping-a systematic review.
\newblock {\em Gait \& postureh}, pages 182--189, 2020.

\bibitem{ekman1997face}
R.~Ekman.
\newblock {\em What the face reveals: Basic and applied studies of spontaneous
  expression using the Facial Action Coding System (FACS)}.
\newblock Oxford University Press, USA, 1997.

\bibitem{Fridlun2014}
A.J. Fridlund.
\newblock Human facial expression: An evolutionary view.
\newblock {\em Academic Press}, 2014.

\bibitem{livingstone2018ryerson}
S.R. Livingstone and F.A. Russo.
\newblock The ryerson audio-visual database of emotional speech and song
  (ravdess): A dynamic, multimodal set of facial and vocal expressions in north
  american english.
\newblock {\em PloS one}, 13(5):e0196391, 2018.
\newblock https://smartlaboratory.org/ravdess/.

\bibitem{ramsay1997functional}
J.~Ramsay and B.W. Silverman.
\newblock {\em Functional data analysis}.
\newblock Springer, 1997.

\bibitem{UllahShahid2013}
S.~Ullah and F.~F. Caroline.
\newblock Applications of functional data analysis: A systematic review.
\newblock {\em BMC medical research methodology}, 2013.

\bibitem{wrobel2018register}
J.~Wrobel.
\newblock Register: Registration for exponential family functional data.
\newblock {\em Journal of Open Source Software}, 3(22):557, 2018.

\bibitem{wrobel2019register}
J.~Wrobel, V.~Zipunnikov, J.~Schrack, and J.~Goldsmith.
\newblock Registration for exponential family functional data.
\newblock {\em Biometrics}, 75(1):48--57, 2019.

\end{thebibliography}
